# PreGSU: A Generalized Traffic Scene Understanding Model for Autonomous Driving based on Pre-trained Graph Attention Network

Yuning Wang, Zhiyuan Liu, Haotian Lin, Junkai Jiang, Shaobing Xu*, Jianqiang Wang*

***Abstract*—Scene understanding, defined as learning, extraction, and representation of interactions among traffic elements, is one of the critical challenges toward high-level autonomous driving (AD). Current scene understanding methods mainly focus on a single concrete task, such as trajectory prediction and risk level evaluation. Although they perform well on specific metrics, the generalization ability is insufficient to adapt to the complexity of real-world traffic and the diversity of downstream demands. In this study, we propose PreGSU, a generalized pre-trained scene understanding model based on graph attention network to learn the universal interaction and reasoning of traffic scenes to support various downstream tasks. After feature engineering and sub-graph module processing, all elements are embedded as nodes to form a dynamic weighted graph. Then, four graph attention layers are applied to learn the relationships among agents and lanes. In the pre-train phase, the understanding model is trained on two self-supervised tasks: Virtual Interaction Force (VIF) modeling and Masked Road Modeling (MRM). Based on the artificial potential field theory, VIF modeling enables PreGSU to capture the agent-to-agent interactions while MRM extracts agent-to-road connections. In the fine-tuning process, the pre-trained parameters are loaded to derive detailed understanding outputs. We conduct validation experiments on three datasets and two downstream tasks, i.e., trajectory prediction in urban scenario and intention recognition in highway scenario, to verify the model's generalization and understanding capabilities. Results show that compared with single-task-driven baselines, PreGSU achieves competitive performance on all datasets and downstream tasks, indicating its potential to be generalized to various scenes and targets. Ablation study shows the effectiveness of pre-train task design.**

## I. Introduction

Autonomous Driving (AD) has made great progress in recent years and is anticipated to significantly enhance traffic efficiency and safety. Measured by the automation-level defined by SAE, many industrial manufacturers have proposed L2+ AD applications, realizing partially automated functions in regular traffic scenes. However, the ability of current systems to cope with complicated interactive scenarios still needs to be proven [1]. Towards high-level AD there are still difficulties.

Scene understanding, defined as the learning, extraction and representation of the interactions in traffic scenarios through raw perception information, is one of the critical challenges for high-level AD [2, 3]. The AD technology framework can be divided into perception, scene understanding, decision-making, and control [4]. Scene understanding involves risk assessment, trajectory prediction, intention recognition, etc. [5], and its role is to reason the inner relationships among the vehicles, road, and other traffic elements so that the decision-making module could have sufficient restrictions and goals to generate appropriate behaviors [6]. With the breakthrough of deep learning (DL) method in recent years, some end-to-end models that directly output control quantities with raw perception inputs were proposed. Most methods incorporate scene understanding as encoders or feature engineering modules within the broader DL network [7]. In summary, whether for hierarchical or end-to-end architecture, scene understanding is an inevitable procedure.

Current scene understanding methods can be roughly divided into two categories, i.e., rule-based models and the data-driven ones. Li et al. surveyed scene risk assessment metrics and concluded that, in specific traffic scenarios, pre-defined models such as time-to-collision and time headway can effectively capture interactions [8]. On the other hand, thanks to the emergence of large-scale AD datasets, DL models are capable of learning interactions among the agents, roadmap, and drivers [9]. Su et al. proposed a surrounding-aware LSTM model to predict vehicle behavior in highway car-following scenarios [10]. Li et al. utilized deep reinforcement learning based on transformers to make lane change strategies [11]. Although these approaches have achieved good performance in specific situations, two main challenges hinder them in high-level AD applications. First, current scene understanding modules lack generalization ability and are only designed for limited tasks and traffic conditions [5]. Second, focusing on detailed downstream tasks such as trajectory prediction or risk level judgment results in overfitting to non-common scenario features. For instance, a model trained by car-following dataset cannot fully understand the interactions in crowded urban scenes, causing improper outputs [12]. In summary, it's hard to strike a balance between performance and generalization ability.

This work is supported by the National Natural Science Foundation of China under Grant 52131201 (the key project), 52372415, and 52221005. This work is also supported by Tsinghua University Toyota Joint Research Center for AI Technology of Automated Vehicle (TTRS 2023-06) (*Corresponding author: Shaobing Xu, Jianqiang Wang.*)

Yuning Wang, Zhiyuan Liu, Junkai Jiang, Shaobing Xu, and Jianqian Wang are with School of Vehicle and Mobility, Tsinghua University, Beijing 100084, China.

Haotian Lin is with the Robotics Institute, Carnegie Mellon University, Pittsburgh, PA 15213, USA.

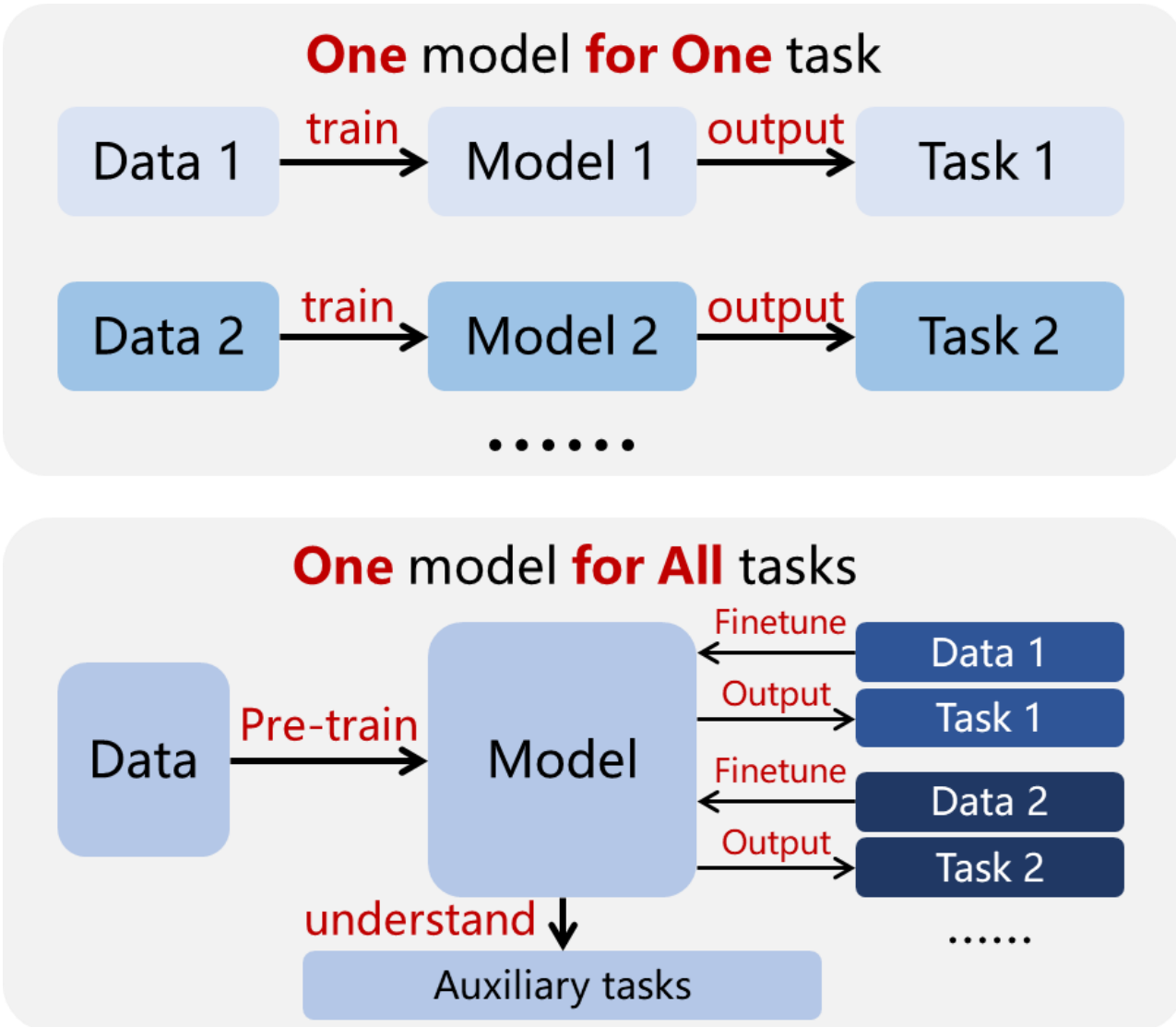

Fig. 1. Comparison of two scene understanding modes

Fig. 1 demonstrates two different scene understanding modes. As summarized above, current scene understanding models adopt a "one for one" paradigm, which designs an independent model based on a single dataset for a specific target task. Under this paradigm, the total model size is large since parameters across various tasks cannot be generalized and reused. For each new task, an entire training process is needed, which is time-consuming. Instead of focusing on specific downstream tasks, a more effective approach is the 'one for all' paradigm, which designs a unified model for multiple tasks. As illustrated in Fig. 1, a general model first learns the basic and interaction representation by pre-training on auxiliary tasks, then conducts small-scale finetuning to finish understanding tasks. Compared with the task-driven mode, this pattern shares a common backbone model, reducing the total parameter size. In addition, the finetuning process only requires relatively small-scale data and fewer epochs for training, reducing the time and data requirements for developing new understanding tasks. Based on the above analysis, this study aims to build a "one for all" universal medium layer for scene understanding, ensuring the adaptation of various succession downstream task modules.

In recent research, graph-based DL methods have been demonstrated to be an effective medium for representing traffic scene understanding, capturing spatiotemporal dependencies [13, 14]. By encoding agents, lanes, and other elements into subgraph nodes and interaction features into edges, a scene can be represented by an interpretable style while also transitive by neural networks, as shown in Fig. 2. In this study, we adopt the dynamic weighted graph as the data structure and graph attention network as the model backbone structure. Apart from the data structure and model backbone, training procedure also plays a vital role in model performance. Pre-train mechanism has been proven to be useful in enhancing the model performance and interpretability in DL applications such as natural language processing and perception [15], which inspires traffic scene understanding. In this research, in contrast with training on the final task directly, we utilize pre-train into the traffic understanding model to increase the understanding performance and model generalization ability.

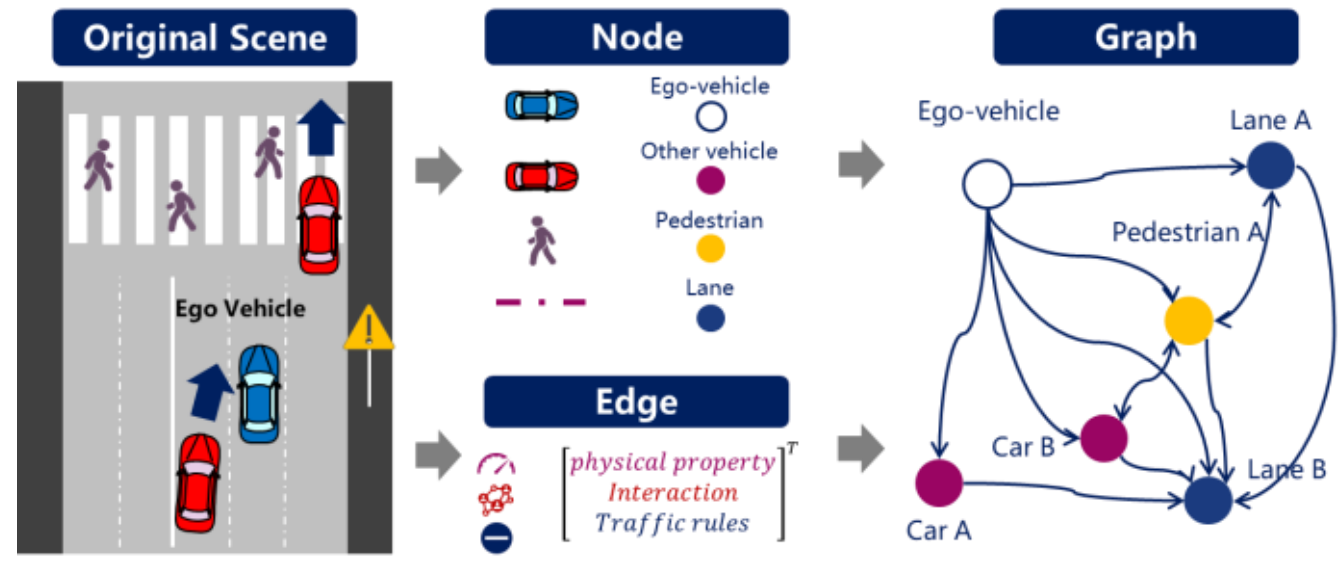

Fig. 2. An illustration of representing scene by graph

In this study, we propose PreGSU, a novel "one for all" scene understanding model based on pre-trained graph attention networks for learning spatial-temporal interactions in traffic scenarios, serving as a universal intermediate layer for various downstream evaluation tasks. A self-supervised graph attention network is constructed based on the two designed pre-train tasks: masked roadmap modeling (MRM) and virtual interaction force (VIF) modeling. The MRM task captures the agent-to-lane relationship, while the novel VIF modeling task based on driving safety field theory enables agent-to-agent reasoning. To illustrate the generalization ability of the proposed model, we validate the performance of PreGSU on two different downstream tasks: multi-modal trajectory prediction in urban scenarios and intention recognition in highway scenarios. The key contributions of this study are summarized as follows:

1) We propose PreGSU, a novel scene understanding framework that adopts the "one for all" paradigm based on pre-trained graph attention network, realizing one model for generalized multiple task understanding.
2) Based on the analysis of traffic interaction mechanisms, we design two corresponding pre-train tasks, which are virtual interaction force modeling and masked road modeling, to learn the fundamental interactions among agents and roadmap for PreGSU.
3) Experiments on specific downstream tasks with different scenarios and output demands demonstrate that PreGSU can realize precise scene understanding, proving good generalization ability.

The remainder of this paper is organized as follows: Section II summarizes the related works about scene understanding methods involving the rule-based and DL-based categories. Section III presents the detailed structure of PreGSU, encompassing the overall pre-training framework, feature engineering, backbone network design, pre-training task design, and training procedures. Experiments and analysis on pre-train and downstream tasks are conducted in Section IV. Section V gives the conclusions and future works.

## II. Related Works

### *A. Rule-based scene understanding models*

Rule-based scene understanding can be divided into four categories. The first category measures the interactions by simple metrics based on physical properties such as TTC, THW, etc. [16]. Some studies use Gaussian distribution to consider the uncertainty in situation awareness [17, 18]. Although these metrics are lucid and explainable, they cannot cope with complicated scenarios where risk exists on multiple dimensions.

The second class utilized a finite state machine that switches among predefined modes and sets the corresponding understanding outputs. Ulbrich et al. used embedded state machines in the Markov process to assess the decision-making under lane-changing scenarios [19]. While the performance is good in specific scenarios, such as a fixed intersection, it cannot deal with uncertain new scenes, leading to poor generalization performance.

The third category establishes the interaction system based on game theory. Bahram et al. proposed a cooperative-driving prediction and planning framework based on dynamic game theory [20]. Li et al. further defined the game theoretic modeling among the driver and the vehicles toward autonomous vehicle control [21]. This category demands manual discretization of driver's behaviors, and computation cost grows exponentially when more participants engage.

Artificial potential field is also a widely used rule-based method for scene understanding. Transferring events into individual sources that generate energy, scenario interaction intensity can be expressed by a quantitative and continuous energy field distribution. Hang et al. used field energy as the risk assessment medium tool to make behavior decisions [22]. Wang et al. proposed Driving Safety Field consisting of three types of various field models [23]. Although field-based methods can effectively reflect the interaction of the scenario, they require complex traverse computation on a densely rasterized grid map, which makes the real-time efficiency not suitable for on-road applications.

Based on the above review, most existing rule-based methods rely deeply on predefined assumptions, which establish their inherently scenario-driven and task-driven nature and result in low generalization ability. In addition, models such as artificial potential field demand complex computation, failing to meet the real-time efficiency demand. However, some of the rule-based methods can effectively reason the interactions in the scenarios, serving as important references for offline modules that do not require real-time efficiency. In this study, we refer to Driving Safety Field, which reflects the human recognition mechanism and interaction pattern well, to design the pretrain tasks for the proposed PreGSU.

### *B. Deep learning scene understanding models*

Due to the size explosion of autonomous driving datasets in the past few years, a lot of deep learning scene understanding models have emerged. Since scene understanding is not an abstract concept, most studies focus on a single scenario or task such as trajectory prediction, which possesses explicit evaluation metrics. In the early stage, sequential models such as Recurrent Neural Network (GRU) and Long-short Term Memory (LSTM). Park et al. proposed an LSTM-based encoder to analyze the fundamental patterns in past trajectories and an LSTM-based decoder to generate future trajectory sequences [24]. Other than sequential models, a Dynamic Bayesian Network is also applied to contain more vehicle dynamics [25, 26]. Since the attention layer was proposed [27], attention-based models including Transformer have become the mainstream backbone model for trajectory prediction, proving to be more effective. Chen et al. utilized a non-autoregressive transformer based on a multi-head attention layer to predict highway scenarios [28]. Tian et al. applied a masked autoencoder mechanism on transformer networks to conduct trajectory prediction in urban cases [29]. Although the prediction accuracy is greatly improved through these methods, the connection to autonomous applications is limited due to a lack of generalization ability.

Some studies break the limitations of single tasks and explore the more macroscopic and generalized task of traffic scene understanding, and the graph-based DL model is a good representation form [30]. Hu et al. proposed UniAD, an integrated framework for scene understanding and planning based on the attention mechanism [31]. Yang et al. proposed a scene understanding module based on a convolutional neural network to measure the importance of the regions [32]. Yu et al. designed a scene-graph data-driven risk assessment method for AD [14]. Monninger et al. established a traffic scene reasoning module based on a graph attention network to model the masked node properties [33]. Though they have extracted scene understanding as an individual module, which is a better understanding pattern compared with single-task models, most studies still validated the performance on a single task and did not change their "one model for one task" essence.

In summary, both current rule-based method and learning-based traffic scene understanding methods are single-scenario or single-task, resulting in low generalization ability, and "one model for all tasks" paradigm meets the demands of real-world application better. In this work, we aim at design a generalized scene understanding method to support various downstream tasks, reducing the transfer learning cost for AD applications.

Pre-train mechanism has been proven to enhance generalization ability and understanding performance in natural language processing and computer vision fields [34, 35]. By first training on simple but general tasks, the model backbone parameters can learn a universal understanding so that with the few-shot fine-tuning process, it can support various specific downstream tasks. In this study, we refer to the learning-based modules to build the backbone model and utilize some rule-based models to design pre-train tasks. By leveraging complementary strengths and mitigating weaknesses, we develop PreGSU, a generalized traffic scene understanding method.

## III. METHODS

### *A. Framework*

The framework of the proposed PreGSU is shown in Fig. 3. The input information consists of a semantic roadmap and the agent history features. In the feature engineering procedure, lanes are expressed by polyline segments to maintain the same input dimension of agent data. Considering the perception error in the dataset, we only consider the lanes and agents within a fixed range, and for those outside the range, a mask is given. The agents include vehicles, pedestrians, bicycles, and other traffic participants. Then to express the vectorized data into the graph structure, the subgraph encoder, which is composed of Multi-Layer Perceptron (MLP), is applied to transfer the lanes and agents into nodes. Subsequently, the nodes are organized into the global graph so that the model could extract the interactions among elements by multiple self-attention and cross-attention layers. Then the scene understanding module is trained on two self-supervised pre-train tasks: Virtual Interaction Force (VIF) modeling and Masked Road Modeling (MRM). VIF modeling task requires the backbone network to predict the virtual interaction forces, defined as the risk energy intensity among agents calculated by Driving Safety Field, through raw traffic scene inputs. MRM requires the model to predict the 50% masked lane topologies through the left known road segments and agent trajectories. Through epochs of pre-train, VIF enables PreGSU to reason the agent-to-agent interactions, and MRM enables the model to learn the interactions among the agents and lanes. After the pre-train procedure, PreGSU has learned the interaction beneath scenarios and has been able to support various kinds of concrete downstream tasks. In this study, we take intention recognition and multi-modal trajectory prediction as two examples of downstream applications of PreGSU, using the performance of these detailed tasks to verify the generalization and understanding ability. For each task, a decoder based on small-scale MLP is utilized to adjust the output format and dimension. Details of PreGSU are shown in the following sections.

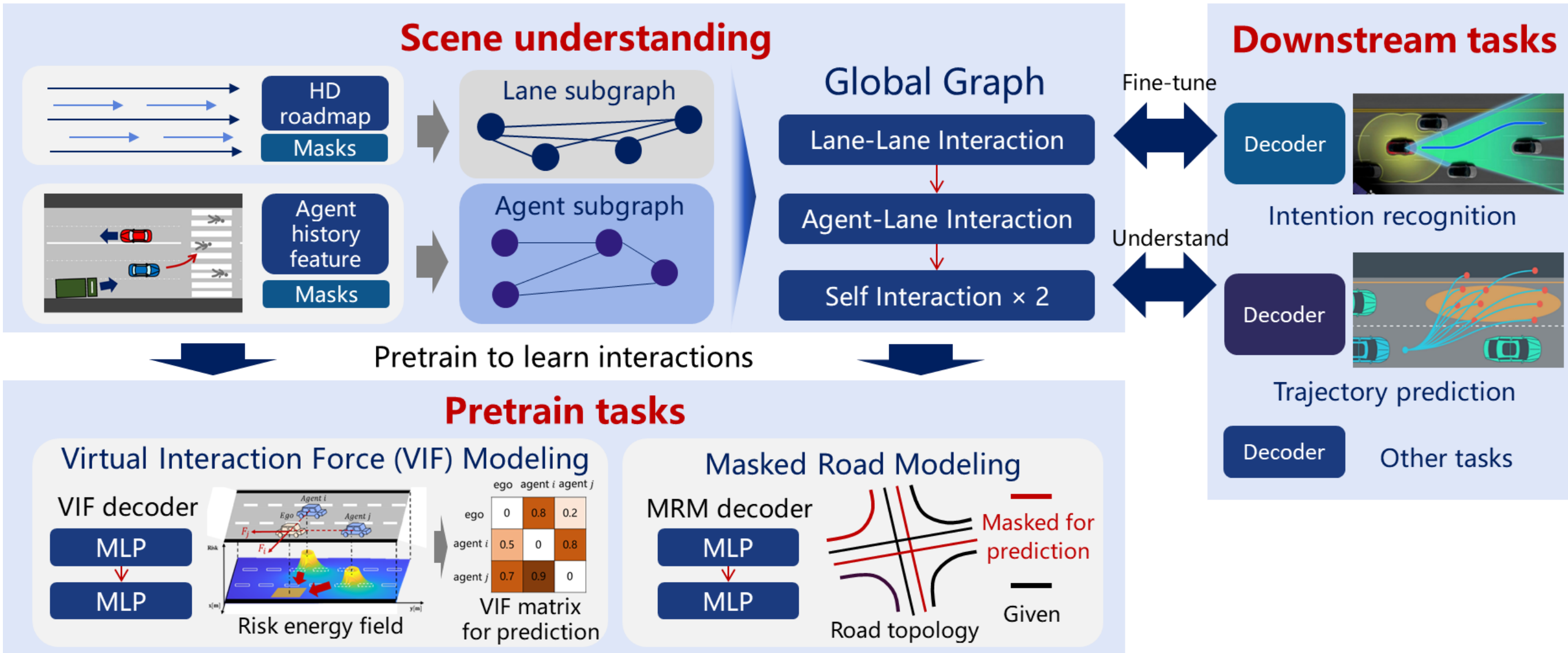

Fig. 3. The framework of PreGSU

*B. Feature Engineering*

From the raw data of complex traffic scenes, we first extract useful features and organize them into a fixed format for model input. Specifically, a piece of traffic scene data usually includes information about the trajectories of agents and information about road topology. Following the vectorized representation [36], we organize the trajectories and road topology into polylines.

*1) Agents*

The trajectory of an agent can be represented by a sequence of states with position, velocity, and attribute information at each timestep. Specifically, the feature vector of an agent $i$ at timestep $t$ can be represented as (1) and (2):

$$A_t^i = [\boldsymbol{start}_t^i, \boldsymbol{end}_t^i, \boldsymbol{v}_t^i, \boldsymbol{a}^i] \tag{1}$$

$$\boldsymbol{v}_t^i = [v_x, v_y] \tag{2}$$

where $\boldsymbol{start}_t^i, \boldsymbol{end}_t^i$ are the coordinates of the start and end points of the agent motion vector at timestep $t$, $\boldsymbol{v}_t^i$ is the velocity vector containing the velocity in the $x$ and $y$ direction, and $\boldsymbol{a}^i$ is the attribute of agent $i$ (i.e., vehicle, pedestrian, cyclist, etc.). The final feature vector $A^i$ of agent $i$ is the concatenate of $A_t^i$ along time $t$, resulting in the shape of $T \times d_a$, where $T$ is the total timesteps of agents' trajectories and $d_a$ is the feature dimension. To be noted that since we have got serial agent position coordinates, the heading angles information have been implicitly included and can be inferred by network.

Following the agent-centric representation, we set the origin of the coordinates at the target agent's current position $\boldsymbol{x_0} \in \boldsymbol{R}^2$ and the positive direction of $x$ −axis on the current heading $\theta_0$. Given a coordinate $\boldsymbol{x}$ of raw data, this transformation can be expressed by (3) and (4):

$$\boldsymbol{x}' = (\boldsymbol{x} - \boldsymbol{x_0}) * \boldsymbol{R} \tag{3}$$

$$\boldsymbol{R} = \begin{bmatrix} \cos\theta_0 & -\sin\theta_0 \\ -\sin\theta_0 & \cos\theta_0 \end{bmatrix} \tag{4}$$

where $\boldsymbol{R}$ is the rotation matrix according to $\theta_0$.

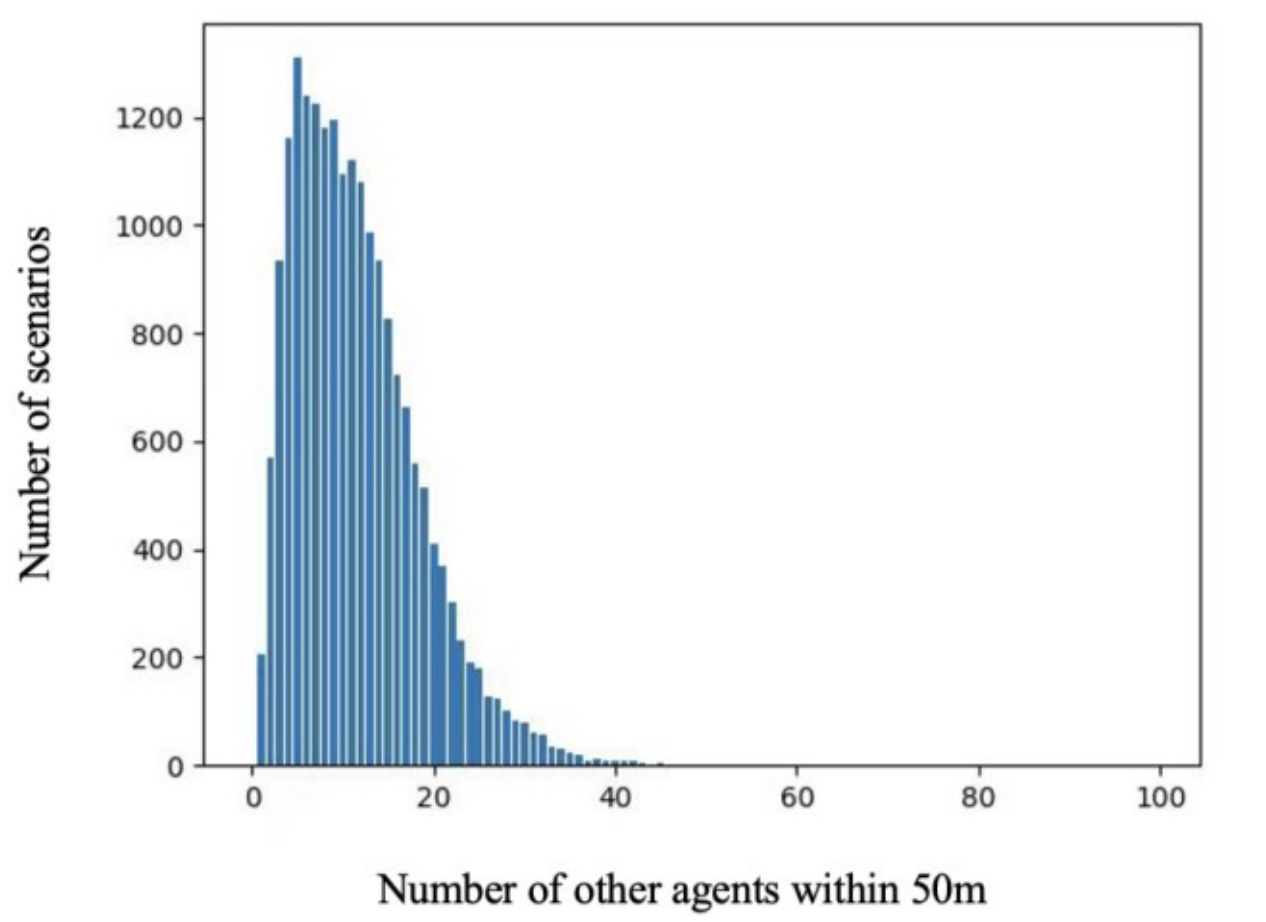

Fig. 4. Distribution of the number of other agents within 50m

One problem for agent feature engineering is that the number of agents could be very large in some scenarios, while only few of them have enough impact on the target agent. Some studies show that considering too many agents may lead to poorer results [37], since the impact of those distant agents on the ego vehicle is subtle. As shown in Fig. 4, most scenarios contain fewer than 20 other agents within 50m. Therefore, we simply filter them by the relative distance to the target agent. Specifically, we sort the relative distance, and the $N_a = 20$ nearest agents are retained. Zero-padding is applied for scenarios that contain fewer than $N_a$ agents. Therefore, all the scenarios will have agent features $\boldsymbol{A} \in \boldsymbol{R}^{N_a \times T \times d_a}$.

*2) Roadmap*

In semantic HD maps, each lane can be represented as a sequence of vectors to approximate its basic geometry. Specifically, the feature vector of the part $j$ of lane $i$ can be represented as (5):

$$M_j^i = [\boldsymbol{start}_j^i, \boldsymbol{end}_j^i, \boldsymbol{a}^i] \tag{5}$$

where $\boldsymbol{start}_j^i$ and $\boldsymbol{end}_j^i$ are the coordinates of the start and end points, and $\boldsymbol{a}^i$ is the binary attribute vector of lane $i$, e.g., whether this segment has a turn direction or controlled by a traffic light. The final feature vector $M^i$ of lane $i$ is the concatenate of $M_j^i$ along the dimension $j$, resulting in the shape of $L \times d_m$, where $L$ is the total number of vectors of a lane, and $d_m$ is the feature dimension.

Similarly, the map features are transformed into the agent-centric coordinate frame. We apply the same filter strategy as agents, where $N_m$ nearest lanes are retained, formulating the map feature $\boldsymbol{M} \in \boldsymbol{R}^{N_m \times L \times d_a}$.

*C. Model Architecture*

As shown in Fig. 3, our scene understanding model consists of two parts: Subgraph Encoder and global graph encoder based on Heterogeneous Attention layers. Subgraph encoder aggregates the history information of each agent and the geometry information of each lane. Then global graph further models the interactions between them. The output of the scene understanding model is a generalized scene representation and can be utilized by different subsequent decoders to achieve variable tasks, such as trajectory prediction and intention recognition.

*1) Subgraph encoder*

In this part, the information of various kinds of agents involving vehicle, pedestrian, bicycle, etc. can be encoded into unified subgraph nodes. Given agent feature input $\boldsymbol{A} \in \boldsymbol{R}^{N_a \times T \times d_a}$ and map feature input $\boldsymbol{M} \in \boldsymbol{R}^{N_m \times L \times d_a}$, the subgraph decoder aggregates the information along dimension $T$ for agent features and dimension $L$ for map features, to obtain polyline-level features $\boldsymbol{A_s} \in \boldsymbol{R}^{N_a \times D}$ and $\boldsymbol{M_s} \in \boldsymbol{R}^{N_m \times D}$ for subsequent interaction modeling, where $D$ is the feature dimension.

Our subgraph encoder is a three-layer PointNet-like polyline encoder [38], as shown in (6):

$$(\boldsymbol{P_j^i})^{(l+1)} = \text{MLP}([\text{MLP}((\boldsymbol{P_j^i})^{(l)}), \varphi(\text{MLP}((\boldsymbol{P^i})^{(l)})]) \tag{6}$$

where $\boldsymbol{P_j^i}$ is the $j^{th}$ segment of polyline $i$, which can be $\boldsymbol{A_t^i}$ or $\boldsymbol{M_j^i}$, MLP is a multilayer perceptron network, $\varphi$ is a max-pooling operator, $l$ represents the $l^{th}$ layer of the subgraph encoder, and $P^i$ denotes all the segments of this polyline. For each layer of the subgraph encoder, the features of each segment are encoded by MLP and interact with each other by concatenating and max pooling. Another MLP is applied for further fusion of each segment. Finally, a max-pooling operation is used to obtain the polyline-level feature, which is conducted by (7).

$$\boldsymbol{P_s} = \varphi((\boldsymbol{P^i})^{(3)}) \tag{7}$$

*2) Global graph encoder*

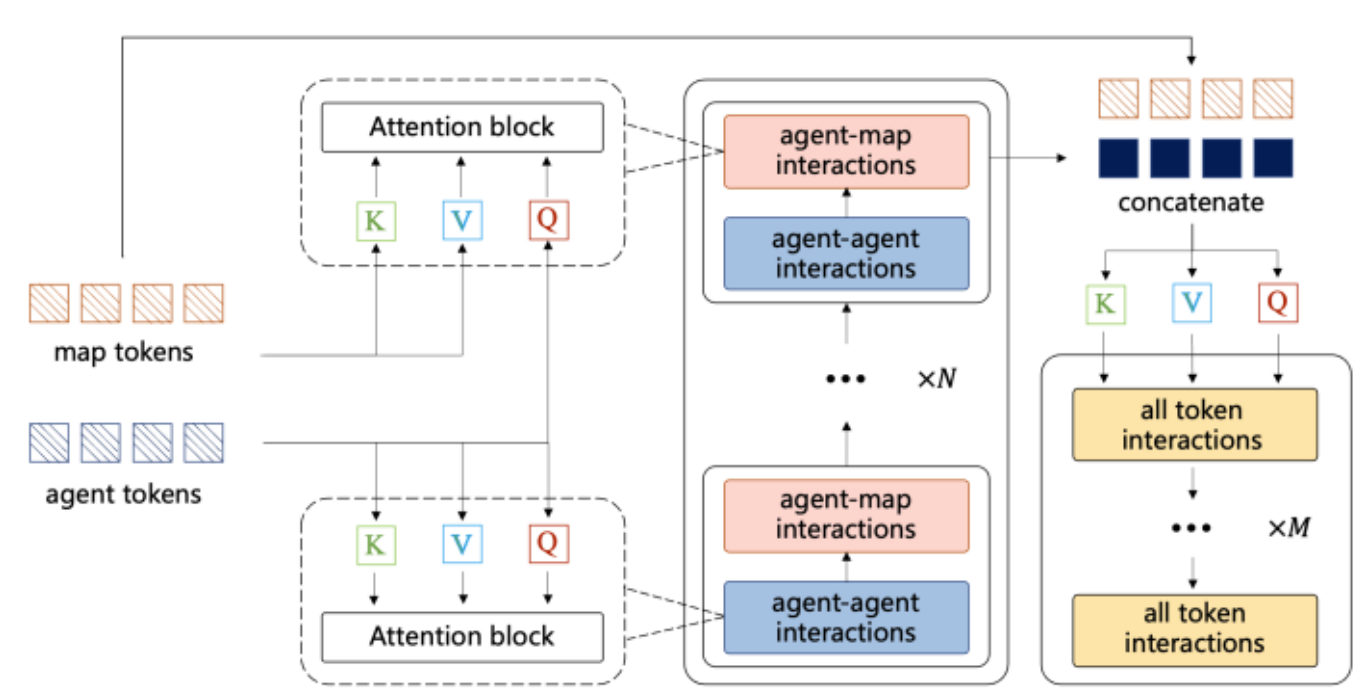

Fig. 5. Hierarchical attention structure of global graph encoder

For each agent, the future behavior is related to the historical state, other agents' behavior, and the road topology. The first item is modeled by subgraph encoder, and the last two still need to be further encoded. Specifically, we use a graph attention network [39] to model the interactions between heterogeneous inputs, i.e., maps and various kinds of agents. Instead of processing all the tokens in the attention layer, we apply the hierarchical structure, which first handles agent interactions as well as map-agent interactions, followed by all token interactions to further encode interaction information, as shown in Fig. 5. This structure benefits from the hierarchical design, reducing the time complexity from $O((N_a + N_m)^2)$ to $O({N_a}^2 + {N_m}^2)$.

For agent interactions, we use the output of the subgraph encoder $A_s$ to formulate query, key, and value, as shown in (8) and (9):

$$\boldsymbol{Q} = \boldsymbol{W_q A_s}, \quad \boldsymbol{K} = \boldsymbol{W_k A_s}, \quad \boldsymbol{V} = \boldsymbol{W_v A_s} \tag{8}$$

$$\widehat{\boldsymbol{A_{s,a}}} = \text{softmax}(\frac{\boldsymbol{QK}^T}{\sqrt{D}})\boldsymbol{V} \tag{9}$$

where $\boldsymbol{W_q}, \boldsymbol{W_k}, \boldsymbol{W_v} \in \boldsymbol{R}^{D\times D}$ are linear layers and $D$ is the feature dimension. Through the attention module, each pair of agents is given an attention score to model interactions between them, then the attention score is used to fuse features from different agent tokens. The result of the attention operation $\widehat{\boldsymbol{A_{s,a}}}$ will be used to update $A_s$ as shown in (10):

$$\boldsymbol{A_s} = \text{MLP}\left(\widehat{\boldsymbol{A_{s,a}}}\right) + \boldsymbol{A_s} \tag{10}$$

where MLP is a multilayer perceptron network, $\boldsymbol{A_s}$ on the right side of the equation is the value before the update and represents the residual connection [40].

For agent-map interactions, we use the output of the subgraph encoder $\boldsymbol{A_s}$ as query and $\boldsymbol{M_s}$ as key and value, as demonstrated in (11) and (12):

$$\boldsymbol{Q} = \boldsymbol{W_q A_s}, \quad \boldsymbol{K} = \boldsymbol{W_k M_s}, \quad \boldsymbol{V} = \boldsymbol{W_v M_s} \tag{11}$$

$$\widehat{\boldsymbol{A_{s,m}}} = \text{softmax}(\frac{\boldsymbol{QK}^T}{\sqrt{D}})\boldsymbol{V} \tag{12}$$

Similar to agent interactions, the attention module calculates the attention intensity between each agent-map pair, representing the allocation of attention to different lanes by each agent, and then fuses information from these lanes. $\widehat{\boldsymbol{A_{s,m}}}$ will be used to update $\boldsymbol{A_s}$ in the same way as described in (10).

The agent interaction and agent-map interaction are repeated $N$ times alternately. After this operation, the updated $\boldsymbol{A_s}$ aggregates preliminary information about agent interactions and road topology. To further fuse the information of all tokens, we introduce the all-token interactions. In this module, $\boldsymbol{C} = [\boldsymbol{A_s}, \boldsymbol{M_s}] \in \boldsymbol{R}^{(N_a+N_m)\times D}$ is used as query, key, and value. Similar to the modules above, the all-token interactions can be calculated as (13), (14), and (15):

$$\boldsymbol{Q} = \boldsymbol{W_q C}, \quad \boldsymbol{K} = \boldsymbol{W_k C}, \quad \boldsymbol{V} = \boldsymbol{W_v C} \tag{13}$$

$$\widehat{\boldsymbol{C}} = \text{softmax}(\frac{\boldsymbol{QK}^T}{\sqrt{D}})\boldsymbol{V} \tag{14}$$

$$\boldsymbol{C} = \text{MLP}\left(\widehat{\boldsymbol{C}}\right) + \boldsymbol{C} \tag{15}$$

All-token interactions are repeated $M$ times, in which all tokens can attend to each other freely to facilitate further feature fusion. Through the designed graph attention network, each token fuses sufficient interaction tokens with other tokens and contains high-level information on agent interactions and road topology, thus can be further used by different downstream decoders.

### *D. Pre-train task design*

The model architecture designed allows the network parameters to have sufficient potential and a reasonable mechanism to learn the ability of scene understanding. Meanwhile, pre-train tasks guide the direction of parameter evolution, which is equally crucial for scene understanding.

In this study, to encompass general and comprehensive features, we start from the mechanism analysis of scenario interaction to design the pre-train tasks. As shown in Fig. 6, analyzing the interaction mechanism, two main factors influence the behavior of the ego vehicle: roadmap restriction and agent restriction [5]. Studies have proven that roadmap topology greatly restricts the vehicle's future motion [41]. Due to the roadmap constraints, the vehicles can only drive along the lanes, enabling all possible vehicle behaviors converge to a limited number of modalities quickly. Since, under most situations, although few intensive agent-to-agent interactions occur, only considering roadmap restriction is sufficient to achieve a good understanding. On the other hand, agent restriction is another crucial factor in understanding the scenario [42]. Because of the agent's influence on each other, vehicles select an appropriate model to drive within the possible lane-restricted trajectories. In summary, in order for the model to adequately learn the scene interaction during the pre-train phase, it is necessary to design pre-train tasks containing roadmap restriction and agent restriction.

In this study, we design the Virtual Interaction Force (VIF) Modelling to learn agent-to-agent reasoning and Masked Roadmap Modelling (MRM) to learn agent-to-lane reasoning.

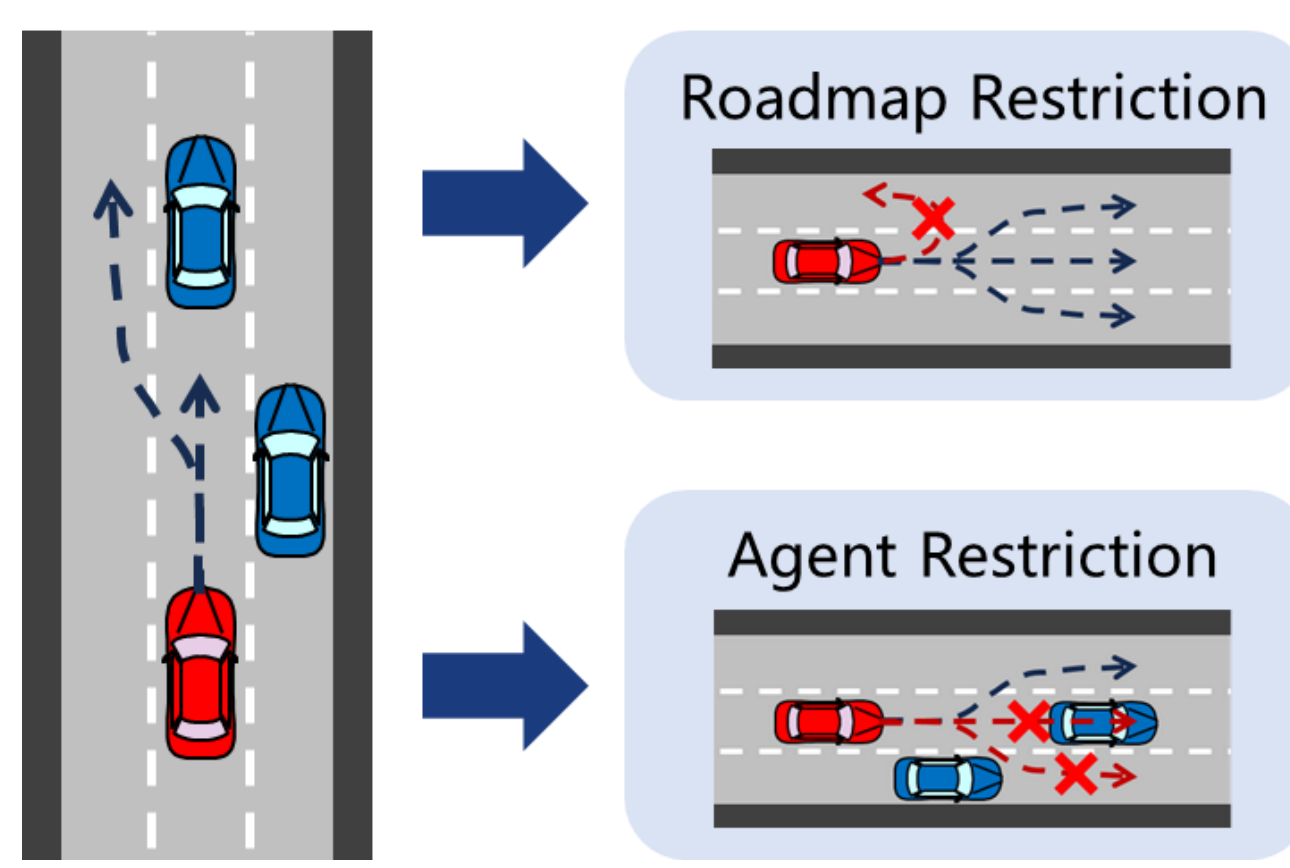

Fig. 6. An analysis of the vehicle interaction mechanism

#### *1) Virtual Interaction Force modeling*

Expressing the interaction among agents is complicated. Raw properties, including positions, velocities, etc., do not contain enough semantic information, while simple metrics such as time-to-collision and time headway only measure the interaction on a specific single dimension. In recent years, field energy theory has been widely used because of its comprehensiveness. Each agent is considered a risk source that radiates energy, and the risk at a certain position in the scene is the sum of energy superimposed by multiple sources. Based on the field theory, Wang et al. proposed Driving Safety Field (DSF) which defined the modeling of the static potential field and dynamic kinetic field [23]. One problem is that computing the field requires multiple high-dimensional iterations on rasterized grid-based scenes thus the real-time performance is poor. For each frame, it takes approximately 0.2 seconds to calculate the field, making it difficult to implement on real-world understanding models. However, the pre-train process does not require real-time efficiency, and in this study, we develop the virtual force concept based on DSF to serve as the supervisory signal of the pre-train task aimed at extracting interactions among agents. According to the definition in [23], the static potential energy $E_{static}$ at position $(x, y)$ generated from an agent at position $(x_a, y_a)$ is calculated as (16), where $M_{eq}$ is the equivalent mass, $r$ is the relative distance, $v_a$ is the agent velocity, $M$ is the vehicle mass, $G$, $a$, $b$, and $c$ are constant coefficients. The dynamic kinetic energy is calculated as (17) to (19), where $\boldsymbol{\Delta v}$ is the relative velocity vector, $\boldsymbol{v}$ is the velocity vector of the target agent, $\boldsymbol{v_a}$ is the velocity vector of the agent, $\boldsymbol{\Delta r}$ is the relative position vector, $k_1$ and $k_2$ are constant coefficients. Then the overall risk energy is computed as (20).

$$E_a^{static} = \frac{GM_{eq}}{r^2} = \frac{GM(a \cdot v_a^c + b)}{(x_a - x)^2 + (y_a - y)^2} \tag{16}$$

$$E_a^{dynamic} = k_1(\boldsymbol{\Delta v})^2 \frac{e^{k_2(\boldsymbol{\Delta v} \cdot \boldsymbol{\Delta r})}}{r} \tag{17}$$

$$\boldsymbol{\Delta v} = \boldsymbol{v} - \boldsymbol{v_a} \tag{18}$$

$$\boldsymbol{\Delta r} = (x - x_a, y - y_a) \tag{19}$$

$$E_a = E_a^{static} + E_a^{dynamic} \tag{20}$$

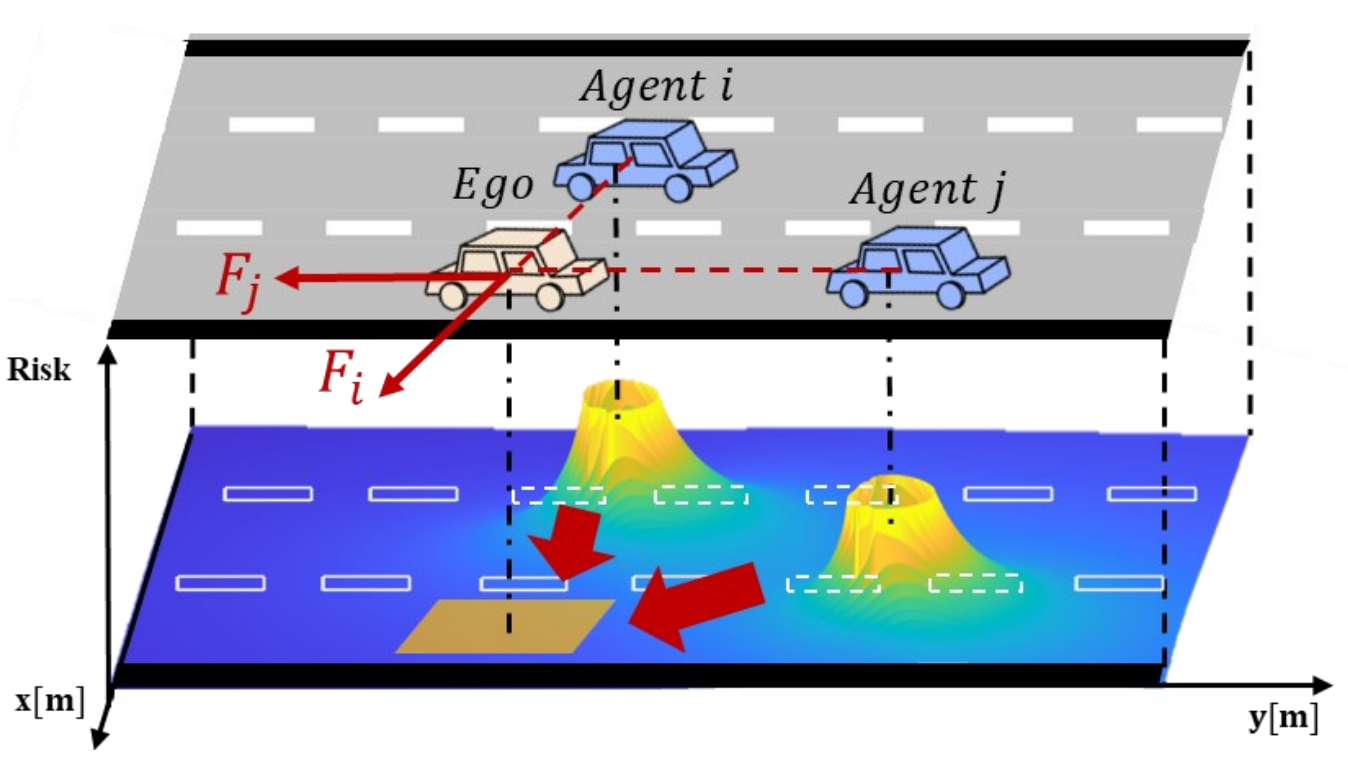

Fig. 7. An illustration of Virtual Interaction Force concept

Based on the field energy, we further apply the VIF concept to transfer the two-dimension field map into an interaction vector. As demonstrated in Fig. 7, considering the area covered by the ego vehicle's bounding box, the virtual force can be decoupled into the components from various risk sources. The force generated from Agent $i$ denoted as $F_i$ is defined as the average energy intensity in the bounding box area $S$, as shown in (21).

$$F_i = \frac{\iint E_i ds}{S} \tag{21}$$

After computing the forces other agents impose on the ego vehicle, normalization in (22) is conducted to show the comparative interaction intensity since, in real traffic, what really matters is the relative influences. Then, organize the forces into the VIF vector as (23), where $N_a$ is the agent number considered in our model as introduced in previous sections. During the pre-train phase, the training target is to fit the ***VIF*** vector for each case with the input of history into information.

$$\overline{F_i} = \frac{F_i - \min_{j \in A} F_j}{\max_{j \in A} F_j - \min_{j \in A} F_j} \tag{22}$$

$$\boldsymbol{VIF} = \left[\overline{F_i} \quad \dots \quad \overline{F_{N_a}}\right] \tag{23}$$

*2) Masked Roadmap Modeling*

As for the MRM pre-training task, we refer to the design in [41]. The semantic roadmap is a graph topology consisting of polyline segments. Each lane segment is connected and neighbored by others, so we can infer the position and shape of a lane through the information about its surroundings. As shown in Fig. 8, the dotted lines represent masked lanes that need to be modeled, and the solid lines denote the known ones.

For each case, the masks are randomly distributed. As for the masking ratio, we refer to the ablation study results in [41] and set it to 50%.

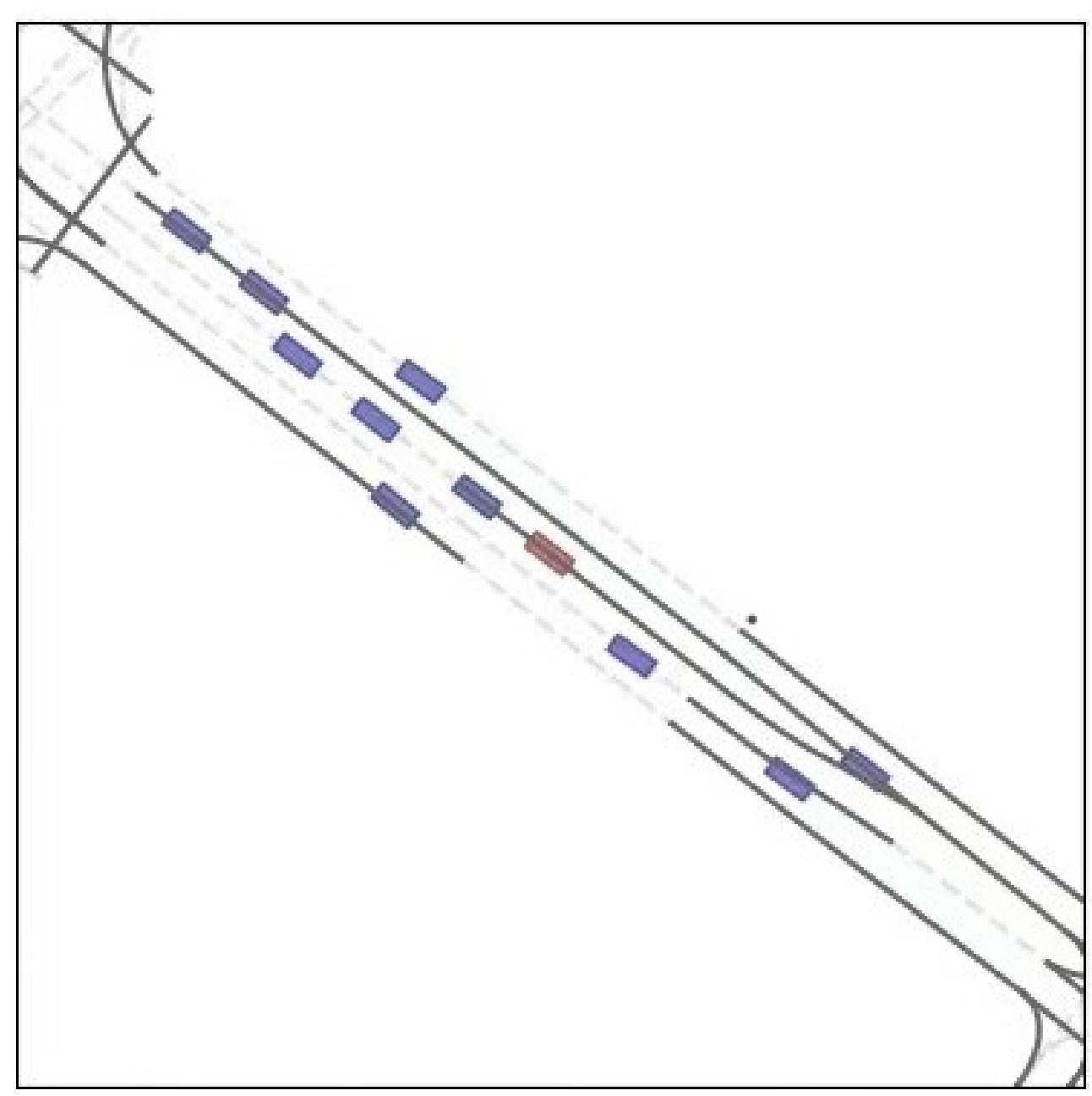

Fig. 8. An illustration of Masked Roadmap Modelling pre-training task

## IV. Implementation and Experiments

In this study, PreGSU is validated on two downstream tasks, which are the multi-modal trajectory prediction in urban environments and the intention recognition in highway environments, and three datasets, which are Argoverse-1, Argoverse-2, and HighD, to validate the scene understanding performance. These two downstream tasks are parallel, and the traditional paradigm is to develop an individual model for each task. In this study, we use one unified backbone model with lightweight fine-tuning to finish various kinds of traffic scene understanding. Diversity in understanding outputs and scenario conditions further test the generalization ability of PreGSU.

The overall procedure of application on a specific task is: pre-training on VIF modeling and MRM, loading the pre-train parameters, and fine-tuning on the tasks.

### *A. Pre-train implementation*

As introduced before, two pre-train tasks are utilized to make the inner parameters of PreGSU learn the spatial-temporal interactions beneath the scenes. One critical issue is how to design the loss function containing two tasks. As for the VIF modeling, we calculate $\mathcal{L}_{VIF}$ as the mean squared errors of the ego vehicle's interaction forces to other agents, as shown in (24)

$$\mathcal{L}_{VIF} = \frac{\sum_{i\in A}\left(\overline{F_i} - \widehat{F_i}\right)^2}{N_a} \tag{24}$$

where $\overline{F_i}$ is the ground truth of VIF, $\widehat{F_i}$ is the predicted force of the ego vehicle to agent $i$, and $A$ is the agent set considered, and $N_a$ is the agent number. To adjust the output format of PreGSU, a simple small-scale decoder is utilized, as shown in Fig. 9.

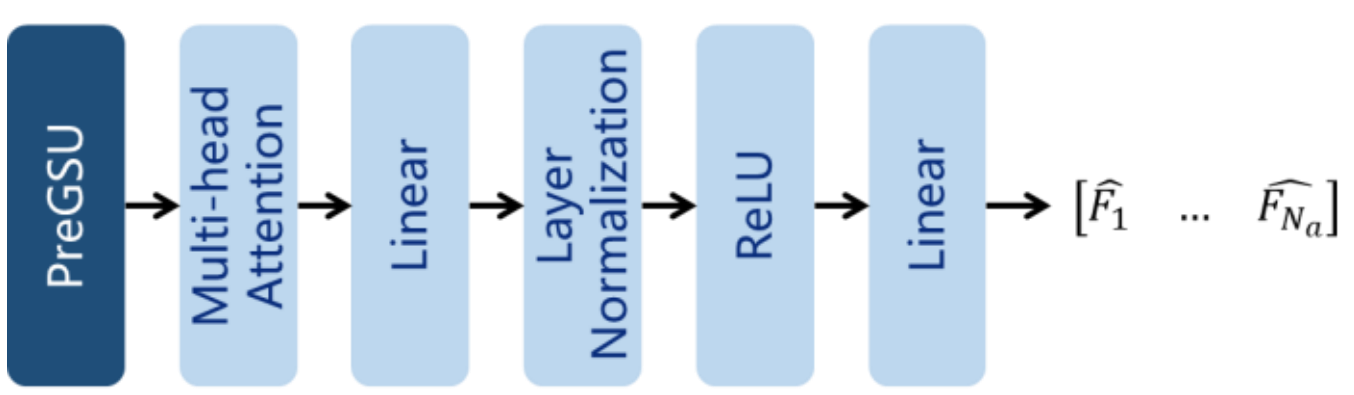

Fig. 9. The decoder for VIF modeling pre-train.

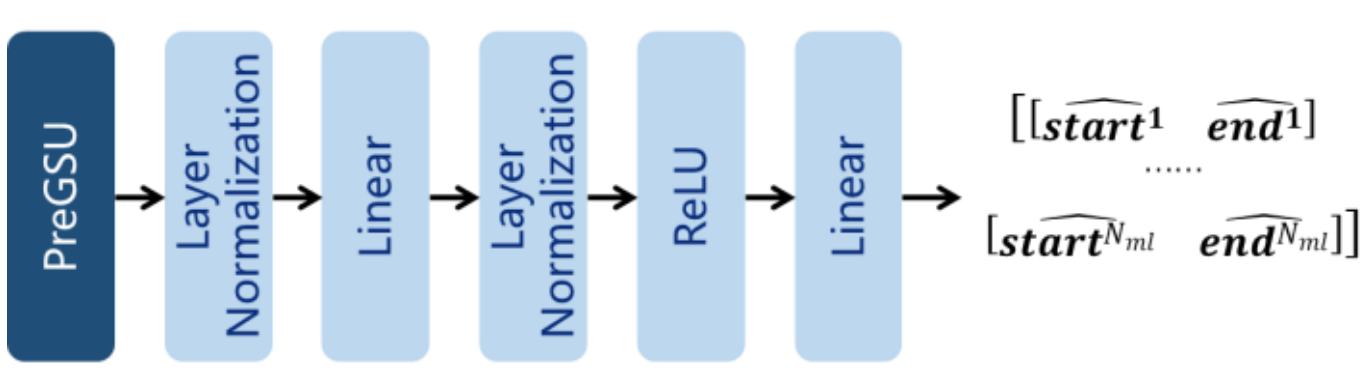

Fig. 10. The decoder for MRM pre-train.

As for the masked lane modeling, we refer to the definition of forecast-MAE [41], using mean squared errors of L2 distance to calculate $\mathcal{L}_{MRM}$, as illustrated in (25)

$$\mathcal{L}_{MRM} = \frac{\sum_{i\in ML}\left\|\boldsymbol{start^i} - \widehat{\boldsymbol{start^i}}\right\| + \left\|\boldsymbol{end^i} - \widehat{\boldsymbol{end^i}}\right\|}{N_{ml}} \tag{25}$$

where $\boldsymbol{start^i}$ and $\boldsymbol{end^i}$ denote the ground truth of start and end points coordinates of lane segments, while $\widehat{\boldsymbol{start^i}}$ and $\widehat{\boldsymbol{end^i}}$ are the predicted coordinate vectors. Similar to VIF modelling, an MLP-based decoder is used to adjust the output dimensions, as illustrated in Fig. 10.

The overall loss is the weighted sum of VIF error and MRM error, as shown in (26). The relative weights $w_{VIF}$ and $w_{MRM}$ are hyperparameters.

$$\mathcal{L}_{Pre} = w_{VIF}\mathcal{L}_{VIF} + w_{MRM}\mathcal{L}_{MRM} \tag{26}$$

Our models are implemented using *Pytorch lightening*. We used ADAM optimizer and a dynamic learning rate following the Cosine Annealing manner. The batch size is set to 64, the drop-out rate is set to 0.1, and the maximum training epoch is 60.

TABLE I
Pre-train Results

| $[w_{VIF}, w_{MRM}]$ | $\mathcal{L}_{VIF}$ | $\mathcal{L}_{MRM}$ |
|---|---|---|
| [1, 1] | 0.0179 | 0.0379 |
| [5, 1] | 0.0175 | 0.0369 |
| [10, 1] | 0.0175 | 0.0329 |
| [40, 1] | 0.0176 | 0.0381 |
| [100, 1] | 0.0178 | 0.0411 |

The training is conducted on a server with an NVIDIA A100 GPU. The pre-train results are shown in Table I. The best weights are selected as $w_{VIF} = 10$ and $w_{MRM} = 1$. The VIF loss reaches 0.0174, and the MRM loss reaches 0.0329. When the weight is too small, the corresponding item cannot generate enough influence on the backward propagation. In contrast, if the weight is too large, the task overfits the data, which can decrease the understanding ability. Therefore, a ratio of 10:1 is the best weight setting, balancing the understanding ability of agents and roadmap interaction. Model parameters are saved as checkpoints, which can be loaded for the downstream tasks.

The parameter size of the backbone understanding model is about 700k, which is lightweight compared with other existing scene understanding models. When coping with detailed downstream tasks, the decoder size does not exceed 20k. The overall model was tested on an NVIDIA GTX 1650Ti GPU, and the results demonstrated that the single-frame inference time is within 5ms, meeting the real-time efficiency requirements for traffic understanding in real-world applications.

### *B. Multi-modal trajectory prediction in urban scenes*

In the trajectory prediction task, we select two large-scale trajectory prediction datasets Argoverse-1 (Argo1) [43] and Argoverse-2 (Argo2) [44] to validate the scene understanding performance. Argoverse-1 provides the trajectory information and HD-map, and we extract the training and validation sets, which contain 205,942 and 38,742 samples, respectively. Each scenario is sampled at 10 Hz and consists of 2 seconds of history and 3 seconds of future. The data format and amount of Argoverse-2 is similar to Argoverse-1, and the main difference is that it contains more challenging interactive cases. The trajectory prediction task is required to predict $K = 6$ future trajectories given the 2-second history and HD-map information.

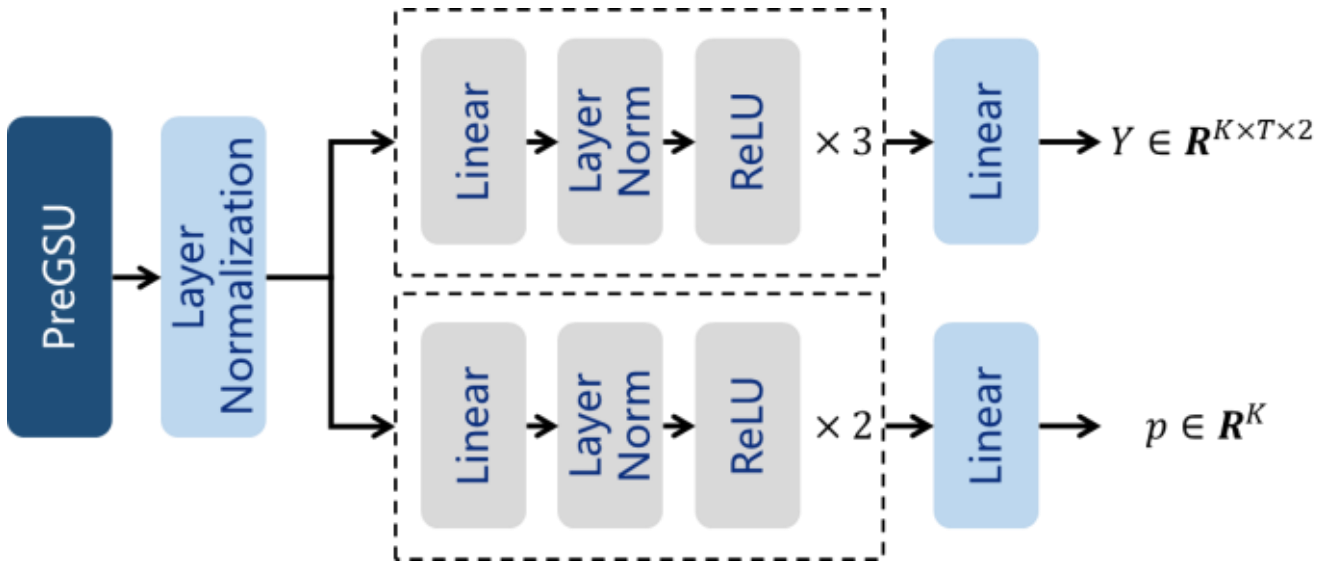

Fig. 11. The decoder for multi-modal trajectory prediction task.

In this task, we use a four-layer MLP to predict the future trajectories $Y \in \boldsymbol{R}^{K \times T \times 2}$, and another three-layer MLP to predict the possibility $p \in \boldsymbol{R}^{K}$ of each trajectory, as shown in Fig. 11. The layers of agent interactions and agent-map interactions $N$ is set to 2, and the layer of all-token interactions $M$ is set to 3. We share the same training settings in the pre-train stage.

For the trajectory regression loss, we only calculate the SmoothL1 loss between the trajectory $Y^{i^*}$ and the ground truth trajectory $\hat{Y}$ with the smallest endpoint distance , as shown in (27) and (28):

$$L_{reg} = SmoothL1(Y^{i^*}, \hat{Y}) \tag{27}$$

$$SmoothL1(Y, \hat{Y}) = \frac{1}{n}\sum_{i=1}^{n} \begin{cases} 0.5(\hat{Y}_i - Y_i)^2 & |\hat{Y}_i - Y_i| < 1 \\ |\hat{Y}_i - Y_i| - 0.5 & |\hat{Y}_i - Y_i| \geq 1 \end{cases} \tag{28}$$

For the classification loss, we apply the Cross-Entropy loss to maximize the possibility $p^{i^*}$ of $Y^{i^*}$, as shown in (29):

$$L_{cls} = -\sum_{i=1}^{K} \mathbb{I}(i = i^*) \log(p^i) \tag{29}$$

where $\mathbb{I}$ is the binary indicator function. The final loss is the sum of the two losses.

Following the official evaluation metrics, we adopt minADE and minFDE to validate the performance. MinADE is the minimum time-average distance between the $K$ predicted trajectories and the ground truth. MinFDE is the minimum end-point distance between the $K$ predicted trajectories and the ground truth. The setting of optimizer, learning rate, and batch size is identical to the pre-train process.

As for validation on Argo1 validation, we select LSTM, vanilla transformer, Efficient Context Graph(ECG) [45], TNT [46], and GOHOME [47], LaneRCNN [48], and QCNet [49] are selected as baselines for comparison. The selected baselines involve multiple types of models. Since PreGSU is based on Graph Attention Networks, two graph-related baselines, ECG and LanRCNN, are also included. To be noted that all selected baselines are small-scale models with fewer than 2M parameters, which creates fair conditions for comparison. The scaling law in AI field implies that as long as the model parameter size is large enough, even simple network architectures can also achieve good performance. However, considering the time efficiency in practical applications, we do not aim to merely stack parameters but rather to maximize parameter efficiency by small-scale powerful models. Therefore, in this study relatively large models such as Prophnet with 15M [50] parameters are not included. The vanilla QCNet also has over 7M parameters, so in this comparison we select the $L_{enc} = 2$ version.

The results of validation on Argoverse-1 datasets are shown in Table II. Among these recent lightweight models, PreGSU achieves the best minADE and competitive minFDE, indicating good scene understanding ability. With 712k model parameters, our model learns traffic scene interaction efficiently. Compared with other graph-related baselines, by applying the proposed two pre-train tasks, the global interaction graph constructed by PreGSU better captures the relationships among traffic elements, resulting in superior performance. Compared with other models specifically developed for trajectory prediction tasks, PreGSU achieves better or comparable performance without directly using trajectory prediction-related metrics for pre-training, further validating its strong generalization capabilities.

TABLE II
MULTI-MODAL TRAJECTORY PREDICTION RESULTS ON ARGO1

| Methods | minADE | minFDE |
|---|---|---|
| LSTM | 0.90 | 1.65 |
| Vanilla Transformer | 0.80 | 1.36 |
| ECG | 0.91 | 1.49 |
| TNT | 0.73 | 1.27 |
| GOHOME | -- | 1.26 |
| LaneRCNN | 0.77 | 1.19 |
| QCNet($L_{enc} = 2$) | 0.73 | 1.27 |
| PreGSU | 0.70 | 1.25 |

TABLE III
MULTI-MODAL TRAJECTORY PREDICTION RESULTS ON ARGO2

| Methods | minADE | minFDE |
|---|---|---|
| LSTM | 0.99 | 1.87 |
| Vanilla Transformer | 0.91 | 1.66 |
| MTR | 0.73 | 1.44 |
| GANet | 0.72 | 1.34 |
| LanRCNN | 0.87 | 1.36 |
| forecastMAE | 0.81 | 1.44 |
| PreGSU | 0.79 | 1.40 |

As for the validation on Argo2, we select LSTM, vanilla transformer, MTR [37], GANet [51], LanRCNN [48], and forecastMAE [41] as baselines. The filtering rule is identical to the experiment on Argo1 that the model parameter size should be lower than 2M. As shown in Table III, the minADE and min FDE are larger than those of Argo1, because the scenarios are more interactive and challenging. Compared with the recent powerful baselines, PreGSU achieves competitive performance on both minADE and minFDE. Through the pre-trained backbone and small-scale fine-tuning, PreGSU learns complicated traffic scene interaction patterns well with a lightweight model. The results of multi-modal trajectory prediction on these two different datasets also prove its adaptability.

To further explore the role of pre-training for scene understanding, we conducted ablation experiments with different pre-training strategies on Argo1, as shown in Table IV. The pretrain mechanism decreases minFDE by 2.27% and minADE by 1.54%. Considering that the selected baselines are quite strong models specially designed for trajectory prediction, it is good for our method which aims at a more generalized usage to achieve such levels of improvement. When only VIF is employed, the model performs even worse than the model without pretrain. A possible reason is that overemphasizing agent interaction may lead to active behavior changes, ignoring the roadmap restriction and making the trajectory similar to that under unstructured environments. When only MRM is applied, the performance is slightly better than the model without pre-train, proving that the MRM strategy helps to understand the road topology. When both strategies are applied, the model understands both agent interactions and road topology better, leading to the best performance in all metrics. In summary, our proposed pretrain strategy has the ability to allow the small-scale model to better understand the scene.

TABLE IV
ABLATION EXPERIMENTS OF PRETRAIN TASK ON TRAJECTORY PREDICTION

| Pre-train Configuration | minADE | minFDE |
|---|---|---|
| None | 0.713 | 1.277 |
| VIF | 0.721 | 1.295 |
| MRM | 0.710 | 1.271 |
| VIF+MRM | 0.702 | 1.248 |

### *C. Intention recognition in highway scenes*

In the intention recognition task, we select the HighD dataset to validate the scene understanding performance. HighD was collected in highway conditions from a bird-eye view, and the behaviors of vehicles can be classified into three kinds, which are driving straight, turning left, and turning right [52]. In the raw dataset, among the original about 11,000 vehicle trajectories, approximately 94.5% of the events are simple car-following. Such an imbalanced distribution will lead to meaningless high-accuracy metrics, failing to understand the high interaction cases. Therefore, we filtered and processed the dataset by retaining all over 6000 left-turn and right-turn trajectories while randomly selecting 6000 straight-driving ones to form a new training and validation set.

The intention recognition can be modeled as a classification problem. Denote $I_a$ as the ground truth of an agent's intention, the output of the model is a predicted distribution $\widehat{p(x)}$ as shown in (30), where $p_l$, $p_s$, and $p_r$ denote the possibility of left, straight, and right accordingly. Then, select the behavior with the highest possibility as the final predicted intention.

$$\widehat{p(x)} = [p_l \quad p_s \quad p_r] \tag{30}$$

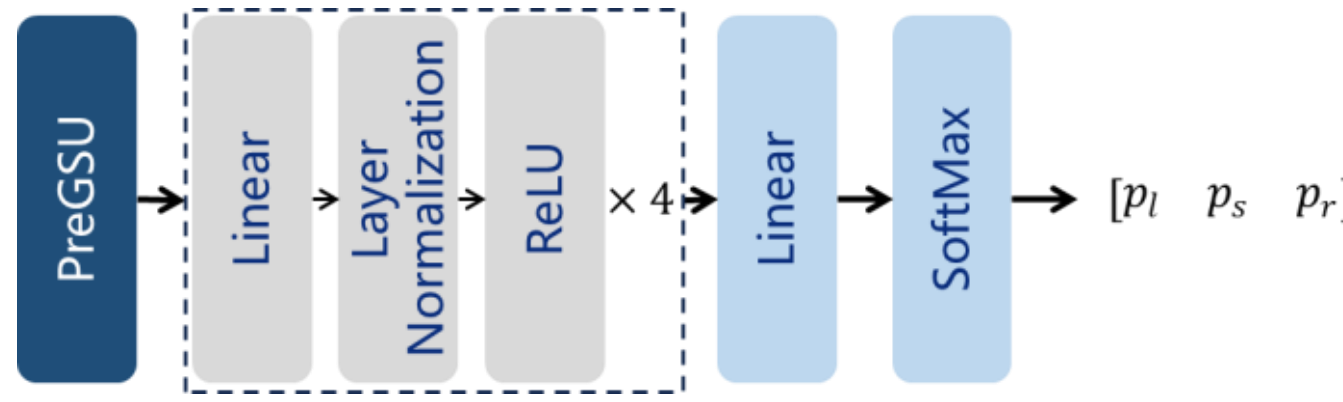

Fig. 12. The decoder for intention recognition task.

The decoder used for PreGSU is a four-layer MLP followed by a SoftMax layer, as shown in Fig. 12. The validation metric is the recognition accuracy, while during the fine-tuning process, the training loss $\mathcal{L}_{IR}$ selects cross entropy as shown in (31) where $p(x)$ is the ground truth binary distribution and $\widehat{p(x)}$ is the prediction intention distribution. The optimizer and learning rate setting is identical to the pre-train process. The batch size is set to 256.

$$\mathcal{L}_{IR} = -\sum_{x \epsilon X} p(x) log\widehat{p(x)} \tag{31}$$

We select GMM [53], LSTM [10], XYST [54], DCIE [55], and GCN [56] as the baselines for comparison. Similar to the trajectory prediction task, this study also includes a graph-related model (Graph Convolutional Network, GCN) in the baselines, further validating the effectiveness of the proposed design. The results of intention recognition accuracy are illustrated in Table V. Our model outperforms all the baselines on the overall accuracy. Specifically, PreGSU exhibits a high accuracy of 98.26% in recognizing left-turn intention, proving a good understanding of interactive scenarios. The accuracy of straight-driving events is also the best. On the other hand, the right-turn accuracy is relatively low, and the possible reason is that sometimes the ground truth is not the most suitable driving according to the traffic rule, while through the pre-train process, the model tends to change to the left lane when both directions are equally drivable. In summary, compared with models specifically developed for intention recognition, PreGSU achieves better performance by learning the intrinsic interaction patterns of traffic scenes, which proves its generalization ability.

TABLE V
INTENTION RECOGNITION ACCURACY RESULTS ON HIGHD

| Methods | Straight | Left | Right | Overall |
|---|---|---|---|---|
| GMM | 87.10% | 93.10% | 92.40% | / |
| LSTM | 93.10% | 94.11% | 88.80% | 93.10% |
| XYST | 90.88% | 91.02% | 92.05% | 91.63% |
| DCIE | 90.50% | 94.80% | 93.90% | 93.07% |
| GCN | 96.80% | 93.90% | 93.50% | 94.73% |
| PreGSU | 94.43% | 98.26% | 91.85% | 95.11% |

The ablation experiment results of the pre-train task design are illustrated in Tabel VI. The pre-train mechanism increases the overall accuracy by 0.96%, and performances of all kinds of intention are improved. To be noted that, as for the intention recognition in highway scenarios, the agent interaction becomes more important since most lane-changing behaviors are caused by the hindrance of other vehicles. Therefore, different from the results of trajectory prediction, only applying the VIF modeling task also enhances the scene understanding ability.

TABLE VI
ABLATION EXPERIMENTS OF PRETRAIN TASKS ON INTENTION RECOGNITION

| Pre-train Configuration | Straight | Left | Right | Overall |
|---|---|---|---|---|
| None | 93.56% | 97.33% | 90.96% | 94.15% |
| VIF | 94.60% | 97.63% | 90.13% | 94.39% |
| MRM | 93.49% | 97.80% | 92.79% | 94.93% |
| VIF+MRM | 94.43% | 98.26% | 91.85% | 95.11% |

To be noted that although in this experiment, multi-modal trajectory prediction and intention recognition are treated as two parallel individual downstream tasks, they can be related and complementary according to the specific application design. For instance, Girase et al. made agent intentions an input dimension for trajectory prediction model to increase precision [57], and Zhang et al. took the reverse way and utilized prior trajectory encoding knowledge to enhance intention recognition accuracy [58]. In addition to the aforementioned two categories, any downstream understanding task can be fine-tuned based on PreGSU. The specific usage and combination depend on the needs and ideas of the user. By learning the intrinsic traffic scene interaction patterns, PreGSU can generalize from the backbone network to various specific understanding outputs and solve multiple tasks with one model.

In summary, validations on three various datasets and two downstream tasks with different scenario conditions and interaction types verify that PreGSU has good generalization ability, learning dynamic complex real-world interaction by a "one for all" paradigm.

### *D. Case visualization*

To further demonstrate the understanding results, some visualization cases are illustrated in Fig. 13 to Fig. 15.

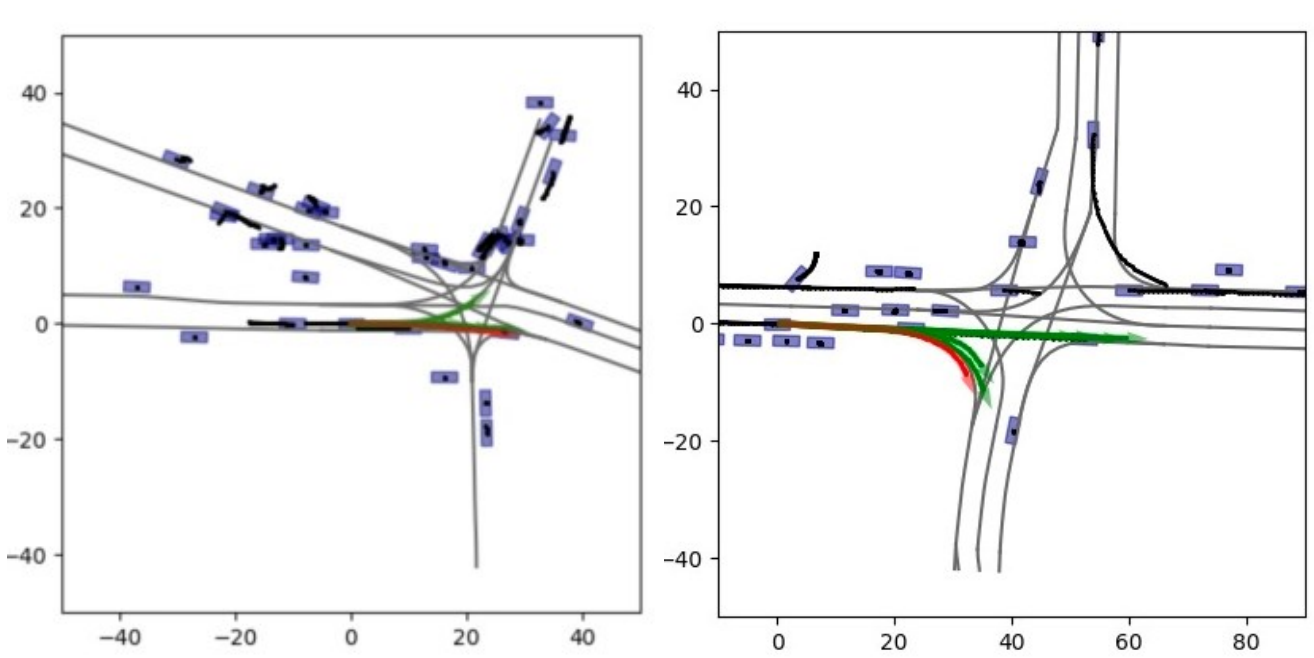

Fig. 13. Visualization cases of vehicle multi-modal trajectory prediction

In Fig. 13 and Fig. 14, the blue marks represent agents, including vehicles, pedestrians, and other traffic participants, the black lines denote agents' history trajectory, the red arrow denotes the future ground truth trajectory of the target agent, and the green arrows denote the predicted multi-modal trajectories. Fig. 13 shows two vehicle prediction cases in the intersection region, in which the proposed method can generate precise future trajectory predictions under complicated surrounding interactions and road topologies. Fig. 14 demonstrates two pedestrian prediction cases, where two pedestrians expect to cross the road. In these two cases, facing surrounding vehicles, the proposed model gives various predicted modals including accelerating passing and turning to yield. The visualizations demonstrate that PreGSU is able to understand urban traffic scenarios well.

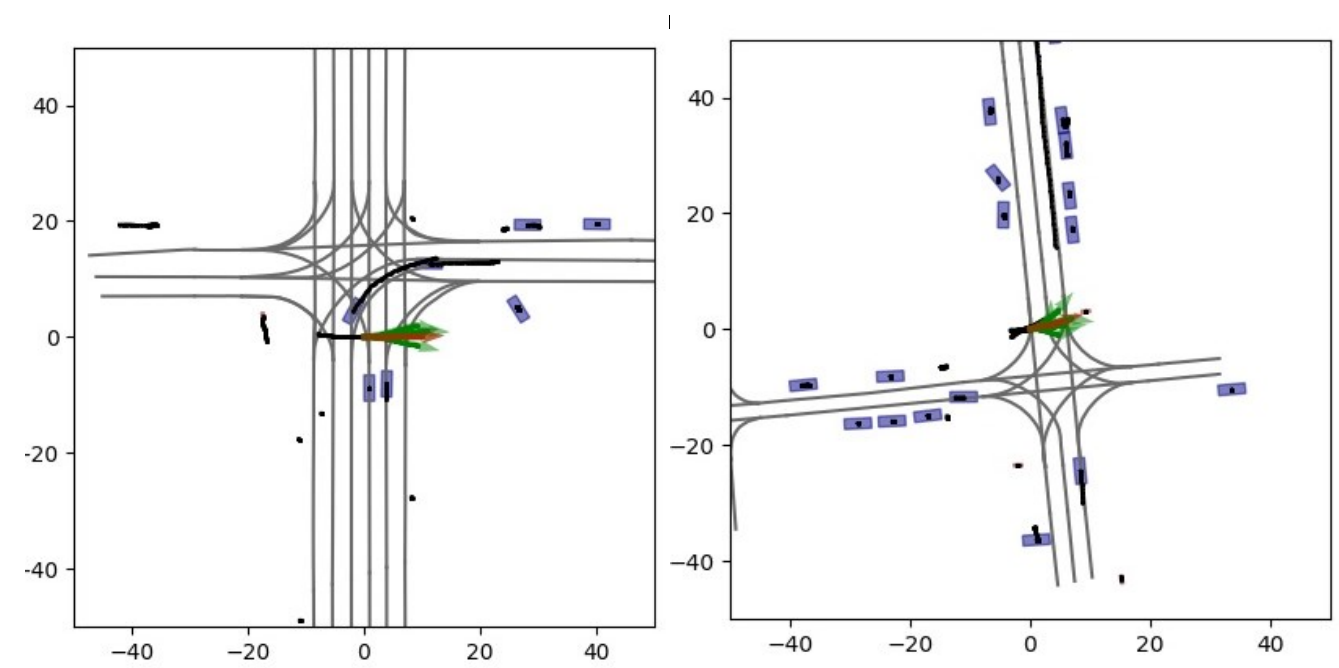

Fig. 14. Visualization cases of pedestrian multi-modal trajectory prediction

Fig. 15 illustrates a highway intention recognition case, where the yellow lines are the road networks, the black marks are the background agents' history trajectory, the blue marks are the target agent's history trajectory, and the green marks are the future ground truth. The output intention is printed at the right top of the figure, which in this case is changing to the left lane. In this case, the preceding vehicle hinders the target agent, so it decides to drive into the left lane, and PreGSU successfully understands the surrounding situation and recognizes its intention correctly. To be noted that although our proposed model can cope with a variety of agent types, under highway scenes only vehicles are permitted to drive in. Therefore, no pedestrian intention recognition case is presented. The visualization results further validate that the proposed PreGSU can understand complicated interactions across various downstream task scenarios and output contents, realizing generalized traffic scene understanding.

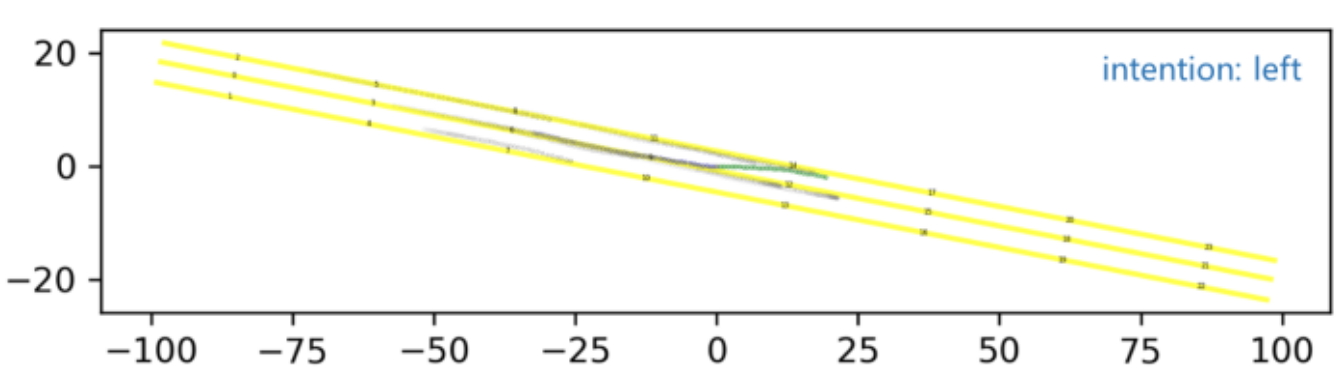

Fig. 15. Visualization cases of intention recognition in highway scenes

Although the analysis above could prove the generalized understanding ability of PreGSU, it still has some limitations. Current datasets include few corner cases, and the agent genres are still limited. For example, corner cases such as a doge suddenly rushes out to cross the road and hinders ego vehicle's behavior are not recorded, which are also critical and real scenarios encountered by drivers. In addition, since U-turns rarely occur in the selected datasets, the understanding ability on them is also limited. Therefore, in the future we will combine corner case scene generation technology to further enhance the generalization ability of the proposed PreGSU.

## V. Conclusions and Future Work

This paper proposed PreGSU, a pre-trained graph scene understanding model, to support various autonomous driving downstream tasks. The self-supervised graph attention network was established based on two designed pre-train tasks: masked roadmap modeling (MRM) and virtual interaction force (VIF) modeling. Experiments on two various downstream tasks, i.e., urban scene trajectory prediction and highway scene intention recognition, and three different datasets validated the scene understanding performance and generalization ability of PreGSU. The results indicated that the proposed model achieves the best performance compared with the baseline models on both tasks. In the urban trajectory prediction, PreGSU achieves competitive minADE and minFDE compared with larger single-task models, while in the intention recognition task the overall accuracy is increased by 2.01%. In addition, ablation studies showed that the pre-train process improves the understanding performance. In summary, PreGSU has a good ability to understand the spatial-temporal interactions in traffic scenes for autonomous driving.

In the future, we plan to explore the decision-making method based on the proposed scene understanding module and validate the performance on real-road tests. We also plan to analyze the transfer cost measured by data amount needed for fine-tuning among various downstream tasks to prepare for future real-traffic applications.

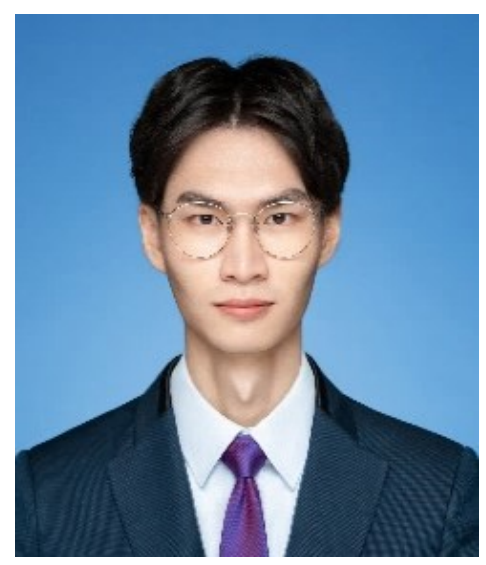

**Yuning Wang** received the bachelor's degree in automotive engineering from School of Vehicle and Mobility, Tsinghua University, Beijing, China, in 2020. He is currently pursuing the Ph.D. degree in mechanical engineering with School of Vehicle and Mobility, Tsinghua University, Beijing, China. His research centered on scene understanding, decision-making and planning, and driving evaluation of intelligent vehicles.

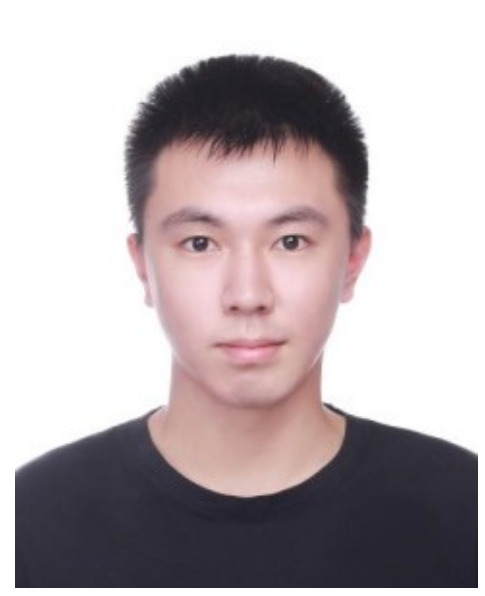

**Zhiyuan Liu** is a senior undergraduate student currently pursuing a bachelor's degree in Mechanics & Vehicle Engineering from Xingjian College, Tsinghua University, Beijing, China. His research interests include trajectory prediction and scene understanding of intelligent vehicles.

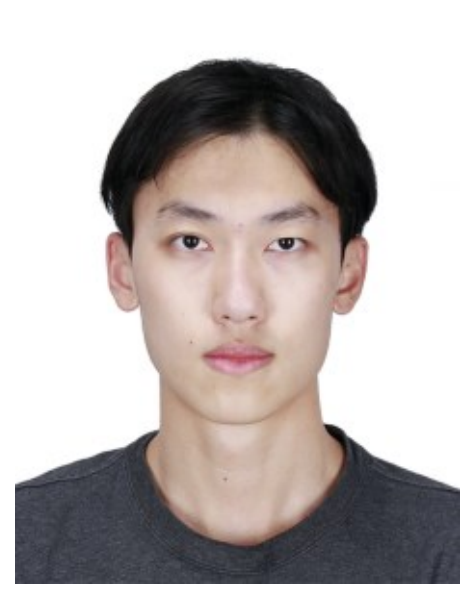

**Haotian Lin** is a senior undergraduate student from Xingjian College, Tsinghua University. His research concentrated on trajectory prediction and decision-making for autonomous vehicles. He is also interested in generative model control for autonomous driving applications.

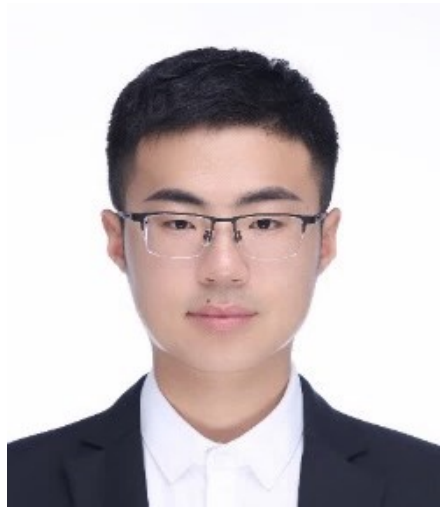

**Junkai Jiang** received the bachelor's degree in automotive engineering from School of Vehicle and Mobility, Tsinghua University, Beijing, in 2021. He is currently pursuing the Ph.D. degree in mechanical engineering with School of Vehicle and Mobility, Tsinghua University, Beijing. His research interests include intelligent and connected vehicles, risk assessment, and decision making.

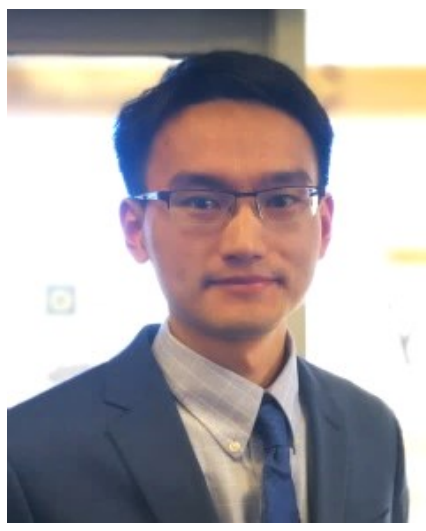

**Shaobing Xu** received his Ph.D. degree in Mechanical Engineering from Tsinghua University, Beijing, China, in 2016. He is currently an assistant professor with the School of Vehicle and Mobility at Tsinghua University, Beijing, China. He was an assistant research scientist and postdoctoral researcher with the Department of Mechanical Engineering and Mcity at the University of Michigan, Ann Arbor. His research focuses on vehicle motion control, decision making, and path planning for autonomous vehicles. He was a recipient of the outstanding Ph.D. dissertation award of Tsinghua University and the Best Paper Award of AVEC 2018 and 2022.

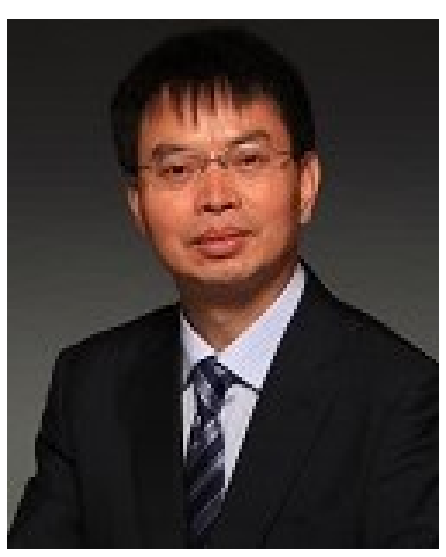

**Jianqiang Wang** received the B. Tech. and M.S. degrees from Jilin University of Technology, Changchun, China, in 1994 and 1997, respectively, and the Ph.D. degree from Jilin University, Changchun, in 2002. He is currently a Professor and the Dean of the School of Vehicle and Mobility, Tsinghua University, Beijing, China.
He has authored over 150 papers and is a co-inventor of over 140 patent applications. He was involved in over 10 sponsored projects. His active research interests include intelligent vehicles, driving assistance systems, and driver behavior. He was a recipient of the Best Paper Award in the 2014 IEEE Intelligent Vehicle Symposium, the Best Paper Award in the 14th ITS Asia Pacific Forum, the Best Paper Award in the 2017 IEEE Intelligent Vehicle Symposium, the Changjiang Scholar Program Professor in 2017, the Distinguished Young Scientists of NSF China in 2016, and the New Century Excellent Talents in 2008.